# Dual-Primal Graph Convolutional Networks

**Federico Monti**[1,2], **Oleksandr Shchur**[3], **Aleksandar Bojchevski**[3], **Or Litany**[4]
**Stephan Günnemann**[3], **Michael M. Bronstein**[1,2,5,6]
[1]USI, Switzerland   [2]Fabula AI, UK   [3]TU Munich, Germany   [4]Tel-Aviv University, Israel
[5]Intel Perceptual Computing, Israel   [6]Imperial College London, UK
{federico.monti,michael.bronstein}@usi.ch   orlitany@mail.tau.ac.il
{bojchevs,shchur,guennemann}@in.tum.de

## Abstract

In recent years, there has been a surge of interest in developing deep learning methods for non-Euclidean structured data such as graphs. In this paper, we propose Dual-Primal Graph CNN, a graph convolutional architecture that alternates convolution-like operations on the graph and its dual. Our approach allows to learn both vertex- and edge features and generalizes the previous graph attention (GAT) model. We provide extensive experimental validation showing state-of-the-art results on a variety of tasks tested on established graph benchmarks, including CORA and Citeseer citation networks as well as MovieLens, Flixter, Douban and Yahoo Music graph-guided recommender systems.

## 1 Introduction

Recently, there has been an increasing interest in developing deep learning architectures for data with non-Euclidean structure, such as graphs or manifolds. Such methods are known under the name *geometric deep learning* [7]. Geometric deep learning has been successfully employed in a broad spectrum of applications, ranging from computer graphics, vision [31, 6, 43], and medicine [37, 48] to chemistry [12, 14] and high energy physics [21].

First formulations of neural networks on graphs [15, 40] constructed learnable information diffusion processes. This approach was improved with modern tools using gated recurrent units [28] and neural message passing [14]. Bruna *et al.* [8, 20] proposed to formulate convolution-like operations in the spectral domain, defined by the eigenvectors of the graph Laplacian. A more efficient class of spectral CNNs are based on filters represented as functions of the Laplacian and expressed in terms of simple operations (scalar- and matrix multiplications, additions, and inversions), including polynomials [10, 26], rational functions [27], and multivariate polynomials [35, 36], thus avoiding explicit eigendecomposition of the Laplacian altogether. Another class of graph CNNs are spatial methods, operating on local neighborhoods on the graph [12, 34, 2, 17, 42]. The current state-of-the-art is [42], which generalizes the attention mechanism [3, 13] to graphs.

The graph attention (GAT) mechanism [42] computes the *importance* of each edge by processing only the features of its incident nodes, thus being "blind" to the neighborhood's behavior. Yet, this ignored information may be useful for better describing the relevance of each vertex, especially in the presence of outliers. To better predict meaningful attention scores and enrich the class of graph filters, we propose to exploit the *dual graph* (also known as the *line-* or *adjoint graph*) for determining neighborhood-aware edge features. The resulting Dual-Primal Graph CNN (DPGCNN) produces richer features and generalizes the GAT mechanism, allowing to achieve better performance on vertex classification, link prediction, and graph-guided matrix completion tasks.

Preprint. Work in progress.

The remainder of this paper is organized as follows: Section 2 reviews Graph CNNs, Section 3 introduces the proposed Dual-Primal GCNN architecture, Section 4 presents experimental results, and Section 5 concludes the paper.

## 2 Graph Convolutional Networks

**Definitions.** Let $\mathcal{G} = \{\mathcal{V}, \mathcal{E}, \mathbf{A}\}$ be a given weighted undirected graph with vertices $\mathcal{V} = \{1, \ldots, n\}$, edges $\mathcal{E} \subseteq \mathcal{V} \times \mathcal{V}$ s.t. $(i,j) \in \mathcal{E}$ iff $(j,i) \in \mathcal{E}$, and edge weights $a_{ij} = a_{ji} \geq 0$ for $(i,j) \in \mathcal{E}$ and zero otherwise. We denote by $\mathcal{N}_i$ a neighborhood of vertex $i$; $\mathcal{N}_i^p$ denotes the $p$-hop neighborhood. The graph structure is represented by the $n \times n$ symmetric adjacency matrix $\mathbf{A} = (a_{ij})$. We define the *normalized graph Laplacian* $\mathbf{\Delta} = \mathbf{I} - \mathbf{D}^{-1/2} \mathbf{A} \mathbf{D}^{-1/2}$, where $\mathbf{D} = \mathrm{diag}(\sum_{j \neq 1} a_{1j}, \ldots, \sum_{j \neq n} a_{nj})$ denotes the degree matrix. In the above setting, the Laplacian is a symmetric matrix admitting an eigendecomposition $\mathbf{\Delta} = \mathbf{\Phi} \mathbf{\Lambda} \mathbf{\Phi}^\top$ with orthonormal eigenvectors $\mathbf{\Phi} = (\phi_1^\top, \ldots, \phi_n^\top)$ and non-negative eigenvalues $0 = \lambda_1 \leq \lambda_2 \leq \ldots \lambda_n$ arranged into a diagonal matrix $\mathbf{\Lambda} = \mathrm{diag}(\lambda_1, \ldots, \lambda_n)$.

We are interested in manipulating functions $f : \mathcal{V} \to \mathbb{R}$ defined on the vertices of the graph, which can be represented as vectors $\mathbf{f} \in \mathbb{R}^n$. The space of such functions is a Hilbert space with the standard inner product $\langle \mathbf{f}, \mathbf{g} \rangle = \mathbf{f}^\top \mathbf{g}$. The eigenvectors of the Laplacian form an orthonormal basis in the aforementioned Hilbert space, allowing a Fourier decomposition of the form $\mathbf{f} = \mathbf{\Phi} \mathbf{\Phi}^\top \mathbf{f}$, where $\hat{\mathbf{f}} = \mathbf{\Phi}^\top \mathbf{f}$ is the *graph Fourier transform* of $\mathbf{f}$. The Laplacian eigenvectors thus play the role of the standard Fourier atoms and the corresponding eigenvalues that of the respective frequencies. Finally, a convolution operation can be defined in the spectral domain by analogy to the Euclidean case as $\mathbf{f} \star \mathbf{g} = \mathbf{\Phi}(\hat{\mathbf{f}} \cdot \hat{\mathbf{g}}) = \mathbf{\Phi}(\mathbf{\Phi}^\top \mathbf{f}) \cdot (\mathbf{\Phi}^\top \mathbf{g})$.

**Spectral graph CNNs.** Bruna *et al.* [8] exploited the above formulation for designing graph convolutional neural networks, in which a basic spectral convolution operation has the following form:

$$\mathbf{f}' = \mathbf{\Phi} \hat{\mathbf{G}} \mathbf{\Phi}^\top \mathbf{f}, \tag{1}$$

where $\hat{\mathbf{G}} = \mathrm{diag}(\hat{g}_1, \ldots, \hat{g}_n)$ is a diagonal matrix of spectral multipliers representing the filter and $\mathbf{f}'$ is the filter output. Here, for the sake of simplicity, we assume a scalar input, though like in classical CNNs, the basic spectral convolution operation (1) can be applied in combination with linear transformations of input and output features, non-linearities, and pooling layers (implemented as graph coarsening). Note that this formulation explicitly assumes the graph to be undirected, since the orthogonal eigendecomposition of the Laplacian requires a symmetric adjacency matrix.

**ChebNet.** Defferrard *et al.* [10] considered the spectral CNN framework with polynomial filters represented in the Chebyshev basis, $\tau_{\boldsymbol{\theta}}(\lambda) = \sum_{j=0}^{p} \theta_j T_j(\lambda)$, where $T_j(\lambda) = 2\lambda T_{j-1}(\lambda) - T_{j-2}(\lambda)$ denotes the Chebyshev polynomial of degree $j$, with $T_1(\lambda) = \lambda$ and $T_0(\lambda) = 1$. A single filter of this form can be efficiently computed by applying powers of the graph Laplacian to the feature vector,

$$\mathbf{f}' = \mathbf{\Phi} \sum_{j=0}^{p} \theta_j T_j(\tilde{\mathbf{\Lambda}}) \mathbf{\Phi}^\top \mathbf{f} = \sum_{j=0}^{p} \theta_j T_j(\tilde{\mathbf{\Delta}}) \mathbf{f}, \tag{2}$$

thus avoiding its eigendecomposition altogether. Here $\tilde{\lambda}$ is a frequency rescaled in $[-1, 1]$, $\tilde{\mathbf{\Delta}} = 2\lambda_n^{-1} \mathbf{\Delta} - \mathbf{I}$ is the rescaled Laplacian with eigenvalues $\tilde{\mathbf{\Lambda}} = 2\lambda_n^{-1} \mathbf{\Lambda} - \mathbf{I}$. The computational complexity thus drops from $\mathcal{O}(n^2)$ as in the case of spectral CNNs to $\mathcal{O}(|\mathcal{E}|)$, and if the graph is sparsely connected (with maximum degree of $\mathcal{O}(1)$), to $\mathcal{O}(n)$. Furthermore, the graph can now be directed, as the framework does not rely on explicit eigendecomposition.

Several follow-up works refined and extended this scheme. Kipf and Welling [25] proposed a simplification of ChebNet (referred to as Graph Convolutional Network or GCN) by limiting the order of the polynomial to $p = 1$ and using a re-normalization of the Laplacian to avoid numerical instability. Levie *et al.* [27] replaced the polynomial filter functions by rational functions based on the complex Cayley transform (CayleyNet), allowing to achieve better spectral resolution of the filters, especially relevant for graphs with communities. Monti *et al.* [36] proposed using graph motifs [33, 5] to create anisotropic kernels (MotifNet). Finally, Monti *et al.* [35] proposed an extension of ChebNet to multiple graphs (Multi-Graph CNN or MGCNN) in the context of graph-guided matrix completion and recommender systems problems.

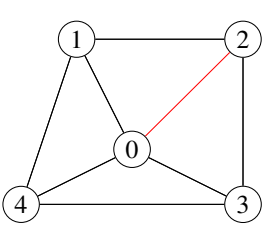 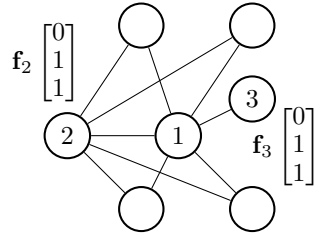

Figure 1: Primal graph (left) and dual graph (right), for the sake of clarity only the ego graph of primal edge $(0, 2)$ was plotted for the dual.

Figure 2: Unlike GAT, DPGCNN is able to distinguish edges $(1, 2)$ and $(1, 3)$, even though $\mathbf{f}_2 = \mathbf{f}_3$.

**Mixture Model Networks (MoNet).** Monti *et al.* [34] proposed a spatial-domain Graph CNN (MoNet) generalizing the notion of 'patches' to graphs. The neighbors of each vertex $i$ are assigned local pseudo-coordinates $\mathbf{u}_{ij} \in \mathbb{R}^d, j \in \mathcal{N}_i$. The analogue of a convolution is then defined as a Gaussian mixture in these coordinates,

$$f'_i = \sum_{m=1}^{M} w_m \sum_{j \in \mathcal{N}_i} \frac{e^{-(\mathbf{u}_{ij}-\boldsymbol{\mu}_m)^\top \boldsymbol{\Sigma}_m^{-1}(\mathbf{u}_{ij}-\boldsymbol{\mu}_m)}}{\sum_{k \in \mathcal{N}_i} e^{-(\mathbf{u}_{ik}-\boldsymbol{\mu}_m)^\top \boldsymbol{\Sigma}_m^{-1}(\mathbf{u}_{ik}-\boldsymbol{\mu}_m)}} f_j, \qquad (3)$$

where $\boldsymbol{\mu}_1, \ldots, \boldsymbol{\mu}_M \in \mathbb{R}^d$ and $\boldsymbol{\Sigma}_1, \ldots, \boldsymbol{\Sigma}_M \in \mathbb{S}_+^d$ are the learnable parameters of the Gaussians. The Gaussians define local weights extracting the local representation of $\mathbf{f}$ around $i$ that can be regarded as a generalization of a 'patch'; the additional learnable parameters $w_1, \ldots, w_M$ correspond to the filter coefficients in classical convolution.

**Graph Attention Networks (GAT).** Veličkovic *et al.* [42] proposed an attention mechanism for directly learning the relevance of each neighbor for the convolution computation. The basic convolution operation with attention has the form:

$$f'_i = \sum_{j \in \mathcal{N}_i^1} \alpha_{ij} f_j, \qquad \alpha_{ij} = \frac{e^{\eta(a([f_i, f_j]))}}{\sum_{k \in \mathcal{N}_i^1} e^{\eta(a([f_i, f_k]))}} \qquad (4)$$

where $\eta$ denotes the Leaky ReLU, and $a([f_i, f_j])$ is some transformation of the concatenated features at vertices $i$ and $j$, implemented in [42] as a fully connected layer. By replicating this process multiple times with different transformations (multiple heads), the authors achieved filters capable of focusing on different classes of vertices in a neighborhoods. We note that GAT can be considered as a particular instance of MoNet (3), where the pseudo-coordinates $\mathbf{u}_{ij}$ are just the features of the nodes $i$ and $j$.

## 3 Learning on dual/primal graphs

One of the drawbacks of graph CNNs described above is that the convolution operations are only applied to vertex features and the underlying domain is considered as fixed. In many situations, this can be a major disadvantage, as the graph can be noisy or only known approximately (e.g. in recommender systems k-NN graphs are typically computed *a priori* based on additional meta-information, which does not necessarily represent the true social relationships existing among users). Furthermore, while vertex features can be very rich, there is no clean mechanism to take advantage of edge features that are more complex than scalars.

The key contribution of our paper is an extension of the graph attention mechanism to edges using the dual graph, whose vertices correspond to the edges of the original graph. As we show in the following, such a formulation allows to address the aforementioned issues and achieves superior performance on a broad range of examples and applications.

**Dual graphs.** Let $\mathcal{G} = (\mathcal{V}, \mathcal{E})$ be a given directed graph, to which we refer as the *primal graph*. The *dual* (also known in graph theory as the *line (di)graph* or *adjoint graph*) of $\mathcal{G}$, denoted by $\tilde{\mathcal{G}} = (\tilde{\mathcal{V}} = \mathcal{E}, \tilde{\mathcal{E}})$, is constructed as follows [19]: each dual vertex $(i, j) \in \tilde{\mathcal{V}}$ corresponds to a primal edge $(i, j) \in \mathcal{E}$, two dual vertices $(i, j), (i', j') \in \tilde{\mathcal{V}}$ are connected by an edge in $\tilde{\mathcal{G}}$ if they share direction and at least an endpoint in $\mathcal{G}$. Figure 1 provides an illustration of this construction for an undirected graph.

We briefly summarize the properties of dual graphs and refer the reader to [19, 18, 16] for additional details. If the primal graph $\mathcal{G}$ is connected, so is its dual $\tilde{\mathcal{G}}$. The dual graph has $\tilde{n} = |\tilde{\mathcal{V}}| = |\mathcal{E}|$ vertices. If $\mathcal{G}$ is undirected, the number of dual edges is $|\tilde{\mathcal{E}}| = \frac{1}{2}\sum_{i=1}^{n} d_i^2 - |\mathcal{E}|$, where $d_i$ denotes the degree of primal vertex $i$. If $\mathcal{G}$ is directed, the dual contains $|\tilde{\mathcal{E}}| = \sum_{i=1}^{n} d_i^{\text{in}} d_i^{\text{out}} - |\mathcal{E}|$ edges [1], where $d^{\text{in}}$ and $d^{\text{out}}$ denote the in- and out-degrees, respectively. The complexity of constructing the dual graph is $\mathcal{O}(|\mathcal{E}|d_{\max}^{out})$, where $d_{\max}^{out} = \max_{i=1,\ldots,n} d_i^{\text{out}}$ is the maximum vertex out-degree in the primal graph (if $\mathcal{G}$ is undirected $d_{\max} = \max_{i=1,\ldots,n} d_i$ should be considered instead of $d_{\max}^{\text{out}}$). While the worst-case complexity is $\mathcal{O}(n^3)$ for fully-connected graphs, for sparsely-connected graphs encountered in practice the cost is linear in $n$.

### 3.1 Dual-Primal GCNN

We propose a *Dual-Primal Graph CNN* (DPGCNN) architecture, which alternates between *dual* and *primal convolutional layers*. The dual convolutional layer applies a GAT on the dual graph to produce features on the edges of the primal graph. These primal edge features are used in the primal convolutional layer to compute attention scores for another GAT, producing primal vertex features. The implementation of both layers are detailed in the following.

**Dual convolution.** Let $\mathbf{F}$ denote the $n \times q$ matrix of input *primal vertex features*, where each row corresponds to a vertex in the primal graph $\mathcal{G}$. The *dual vertex features* (or equivalently, *primal edge features*) $\tilde{\mathbf{f}}_{ij} = [\mathbf{f}_i, \mathbf{f}_j]$ are constructed by concatenating the respective primal vertex features (row vectors $\mathbf{f}_i, \mathbf{f}_j$), for each $(i,j) \in \mathcal{E}$. We denote by $\tilde{\mathbf{F}}$ the $\tilde{n} \times 2q$ matrix of all the dual vertex features arranged row-wise. To avoid ambiguity, for undirected graphs we construct two nodes in the dual for every undirected edge $\{i,j\}$, namely $(i,j), (j,i) \in \tilde{\mathcal{V}}$, and we connect a dual node $(i,j)$ to all the nodes corresponding to edges pointing to $i$ or departing from $j$. This avoids establishing an order among vertices which otherwise would be required to define edge features (if one edge $(i,j)$ is represented by one dual node, one needs to define whether the features of $i$ or $j$ comes first in the concatenation).

Applying GAT on the dual graph $\tilde{\mathcal{G}}$ with features $\tilde{\mathbf{F}}$ has the form

$$\tilde{\mathbf{f}}'_{ij} = \xi_{\text{d}}\left(\sum_{r \in \mathcal{N}_i} \tilde{\alpha}_{ij,ir}\tilde{\mathbf{f}}_{ir}\tilde{\mathbf{W}} + \sum_{t \in \mathcal{N}_j} \tilde{\alpha}_{ij,tj}\tilde{\mathbf{f}}_{tj}\tilde{\mathbf{W}}\right), \tag{5}$$

$$\tilde{\alpha}_{ij,ik} = \frac{e^{\eta(\tilde{a}([\tilde{\mathbf{f}}_{ij}\tilde{\mathbf{W}}, \tilde{\mathbf{f}}_{ik}\tilde{\mathbf{W}}]))}}{\sum_{r \in \mathcal{N}_i} e^{\eta(\tilde{a}([\tilde{\mathbf{f}}_{ij}\tilde{\mathbf{W}}, \tilde{\mathbf{f}}_{ir}\tilde{\mathbf{W}}]))} + \sum_{t \in \mathcal{N}_j} e^{\eta(\tilde{a}([\tilde{\mathbf{f}}_{ij}\tilde{\mathbf{W}}, \tilde{\mathbf{f}}_{tj}\tilde{\mathbf{W}}]))}} \tag{6}$$

where $\tilde{\mathbf{f}}'_{ij}$ denotes the $\tilde{q}$-dimensional output feature at dual vertex (equivalently, primal edge) $(i,j)$, $\tilde{\alpha}_{ij,ik}$ are the dual attention scores, $\tilde{\mathbf{W}}$ is a $2q \times \tilde{q}$ matrix of learnable weights, $\tilde{a}$ is a fully connected layer mapping $2\tilde{q}$-dimensional input into a scalar output, $\xi_{\text{d}}$ is the dual layer activation function (typically, a ReLU) and $\eta$ is the Leaky ReLU.

The dual convolution on $\tilde{\mathcal{G}}$ is equivalent to exchanging information across primal edges which share common directions. In particular, primal edge $(i,j)$ exchanges information only with edges that income to primal vertex $i$ or outgo from from primal vertex $j$. This additional diffusion naturally allows to better characterize the behavior of each single connection, as every edge $(i,j)$ is now not only represented by the features associated with the corresponding incident vertices but also by an aggregated representation of all the edges that present common spreading patterns (i.e., that bring information to $i$ or spread information from $j$). This naturally allows to predict better attention scores. Note that such description could not be achieved with two GAT layers as our aggregation step in equations (5)-(6) depends on dual connectivity plus the concatenation of the features of incident vertices and not on the features of the single vertices themselves. For a concrete example, consider Figure 2. Since vertices 2 and 3 have the same attribute vectors ($\mathbf{f}_2 = \mathbf{f}_3$), primal GAT will produce the same attention scores for edges $(1,2)$ and $(1,3)$. In contrast, dual GAT (our dual convolutional layer) is able to differentiate between them by aggregating different edge features for the two different edges.

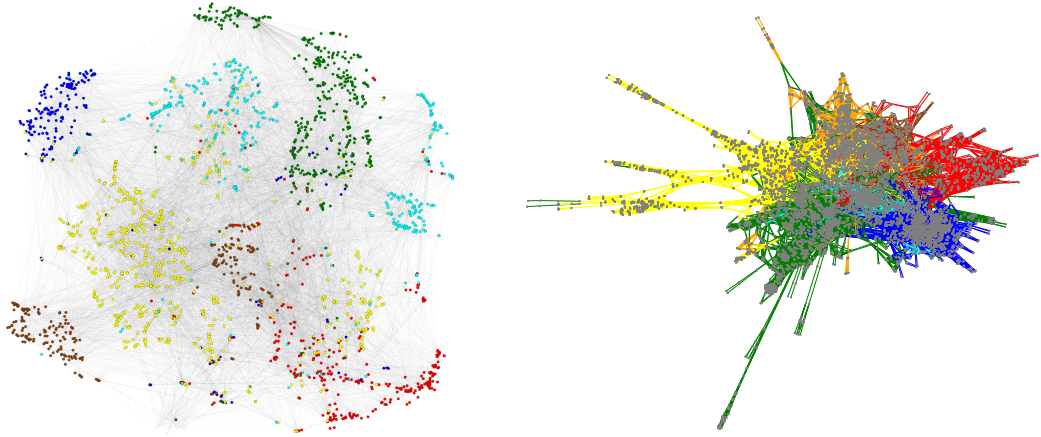

Figure 3: Left: primal graph of the CORA citation network. Vertex colors code the groundtruth classes. Vertex positions are the learned primal vertex features, mapped to the plane using tSNE. Edge thickness represents the edge attention scores. Right: dual graph of CORA. Edge colors represent the groundtruth classes.

**Primal convolution.** The convolution on the primal graph is applied using a GAT on the primal vertex features $\mathbf{F}$. The key difference compared to the simple GAT (4) is that primal attention scores are computed using the dual vertex features $\tilde{\mathbf{F}}'$ produced by the dual convolution,

$$\mathbf{f}'_i = \xi_{\mathrm{p}} \left( \sum_{r \in \mathcal{N}_i} \alpha_{ij} \mathbf{f}_i \mathbf{W} \right), \qquad \alpha_{ij} = \frac{e^{\eta(a(\tilde{\mathbf{f}}'_{ij}))}}{\sum_{k \in \mathcal{N}_i} e^{\eta(a(\tilde{\mathbf{f}}'_{ik}))}}, \qquad (7)$$

where $\mathbf{f}'_i$ denotes the $q'$-dimensional output features at primal vertex $i$, $\alpha_{ij}$ are the primal attention scores, $\xi_{\mathrm{p}}$ is the primal layer activation function, $\mathbf{W}$ is a $q \times q'$ matrix of learnable weights, and $a$ is a fully connected layer mapping $\tilde{q}$-dimensional input into a scalar output.

### 3.2 Architecture Variants

**GAT as an instance of DPGCNN.** The primal and dual convolutional layers can be used as building blocks of graph CNN architectures. The dual convolution precedes one or more primal convolutional layers; multiple layers can be used to obtain deep neural networks. GAT is a particular setting of DPGCNN obtained by setting the dual attention scores $\tilde{\alpha}_{ij,ik} = 0$ for $k \neq j$ and 1 otherwise.

**DPGCNN with polynomial filters.** Convolution with polynomials of normalized adjacency matrices can be obtained with DPGCNN by computing different attention scores for different orders:

$$\mathbf{f}'_i = \xi \left( \sum_{l=0}^{p} \mathbf{f}_i^{(l)} \mathbf{\Theta}_l \right), \qquad (8)$$

$$\mathbf{f}_i^{(k)} = \sum_{j \in \mathcal{N}_i} \alpha_{ij}^{(k)} \mathbf{f}_j^{(k-1)}, \quad \mathbf{f}_i^{(0)} = \mathbf{f}_i \qquad (9)$$

where $\mathbf{\Theta}_l$ denote polynomial coefficients and the recursive definition is similar to [36]. $\alpha_{ij}^{(k)}$ is obtained as described in Eq. 6 or 7 if operating on the primal or dual respectively. Such an approach can be exploited in Eq. 5 or 7 to further enrich the filters on both primal and dual graphs.

**General edge features.** Finally, we have so far assumed for simplicity that the edge features are derived from vertex features, $\mathbf{f}_{ij} = [\mathbf{f}_i, \mathbf{f}_j]$. Our framework naturally allows to apply dual convolutional layers to arbitrary edge features, both vector- or scalar-valued.

# 4 Experiments

## 4.1 Citation Networks

**Vertex Classification.** The first task we consider is a semi-supervised (transductive) learning problem on two standard citation network benchmark datasets (CORA and Citeseer [41]), following the extablished experimental setup used in [46, 26, 34, 42]. The vertices of the citation graph are scientific papers and edges are citations; the graph is assumed to be undirected (i.e., there is an edge $(i,j)$ if $i$ cites $j$ or vice versa). Each vertex is represented by a descriptor capturing the content of the respective paper. The task is to classify each vertex in the graph according to its publication field. CORA contains 2708 vertices, 5429 edges, 7 different categories and 1433 binary features per vertex; Citeseer contains 3327 vertices, 4732 edges, 6 different classes and 3703 features per vertex.

For each dataset we repeat verbatim the experiments presented in [26, 34, 42]. As training set, we use 140 vertices sampled from CORA and 120 from Citeseer, using the split from [46]. We use the architecture of [42] (i.e. 2 convolutional layers, 8 heads for first layer with 8 features as output per head, 1 head for the second layer with # classes as output) with one head and 32 features as output when convolving on the dual graph. Training settings, including dropout, weight decay and learning rate are as in [42]. Table 1 summarizes the results, averaged over 100 runs to account for different random initializations. DPGCNN beats all the competing architectures, albeit by a small margin.

Table 1: Vertex classification accuracy on CORA and Citeseer citation networks, averaged over 100 runs.

| Method | Cora | Citeseer |
|---|---|---|
| MLP | 51.1% | 46.5% |
| ManiReg [4] | 59.5% | 60.1% |
| SemiEmb [44] | 59.0% | 59.6% |
| LP [47] | 68.0% | 45.3% |
| DeepWalk [38] | 67.2% | 43.2% |
| ICA [29] | 75.1% | 69.1% |
| Planetoid [46] | 75.7% | 64.7% |
| ChebNet [10] | 81.2% | 69.8% |
| GCN [26] | 81.5% | 70.3% |
| MoNet [34] | 81.7 $\pm$ 0.5% | – |
| GAT [42] | 83.0 $\pm$ 0.7% | 72.5 $\pm$ 0.7% |
| **DPGCNN** | **83.3 $\pm$ 0.5%** | **72.6 $\pm$ 0.8%** |

We further reproduced a different setting of the same experiment reported in [27], in which 500 vertices were sampled from CORA for training[1]. We used an architecture with two convolutional layers and 16 features as output from convolutional layers on both primal and dual graphs (realized as described in equation (8) with monomial bases and attention). Dual convolution was applied only in the second layer, in order to reduce the overall number of parameters. Attention with one head was used on both primal and dual to provide a fair comparison with CayleyNet [27]. Dropout, weight decay and learning rate were as in [27]. We compare our results to CayleyNet [27] and GAT [42] with the same architecture, using polynomial filters of different order (GAT with polynomial filters was implemented according to equation 8, using primal graph only). Table 2 reports the performance averaged on 20 runs. Our DPGCNN beats the competing architectures.

**Link Direction Prediction.** The second task we address is to predict the direction of links, which we cast as a semi-supervised classification problem on the dual graph. For this experiment, we used a directed version of the CORA graph [36], of which we took a subset containing 1118 vertices (each represented by a 8710-dimensional feature vector) and 4155 directed edges. All the edges were turned into undirected; given an undirected edge $\{i,j\}$, the goal was to predict the direction of the original edge. The dual graph contains every edge in the two possible directions i.e. $(i,j), (j,i) \in \tilde{\mathcal{V}}$, link prediction in the primal is thus a binary classification problem on the dual. 10% of edges' directions were used for training, 10% for validation, and 10% for testing.

---
[1]Training/validation/test indices together with CayleyNet performance have been obtained by the authors of the paper, scaled unnormalized laplacian has been used for the reported CayleyNet's accuracies.

Table 2: Number of parameters / Vertex classification accuracy on CORA citation network (500 training samples) using polynomial filters of different order.

| Order $p$ | CayleyNet | GAT | **DPGCNN** |
|---|---|---|---|
| 1 | 46K / 88.1 $\pm$ 0.6% | 46K / 88.65 $\pm$ 0.58% | 47K / **88.92** $\pm$ 0.51% |
| 2 | 69K / 88.0 $\pm$ 0.5% | 69K / 88.00 $\pm$ 0.39% | 71K / **88.22** $\pm$ 0.41% |
| 3 | 92K / 87.6 $\pm$ 0.6% | 92K / 87.54 $\pm$ 0.52% | 95K / **87.69** $\pm$ 0.42% |
| 4 | 115K / 86.4 $\pm$ 0.8% | 115K / 87.06 $\pm$ 0.42% | 118K / **87.30** $\pm$ 0.50% |
| 5 | 138K / 86.5 $\pm$ 0.8% | 138K / 86.67 $\pm$ 0.52% | 142K / **86.68** $\pm$ 0.74% |
| 6 | 161K / 86.7 $\pm$ 0.7% | 161K / 86.38 $\pm$ 0.57% | 165K / **86.50** $\pm$ 0.61% |

Table 3: Link direction prediction accuracy on directed CORA, averaged over 100 runs.

| Method | Accuracy | #Param |
|---|---|---|
| Dual GAT | 72.94 $\pm$ 1.12% | 140K |
| Primal GAT | 74.95 $\pm$ 1.38% | 140K |
| **DPGCNN** | **76.45** $\pm$ 1.07% | 142K |

Three different architectures were tested: GAT operating only on the primal graph (Primal GAT), GAT operating only on the dual graph (Dual GAT), and a DPGCNN operating on both. Three convolutional layers and a final fully connected layer followed by *softmax* were used in all the three architectures. The final FC layer was applied (i) on the concatenation of the features of nodes $i$ and $j$ for Primal GAT since outputs only node features, (ii) on the features of edge $(i, j)$ for Dual GAT, and (iii) on the concatenation of the two for DPGCNN. In Dual GAT, we had an additional initial dimensionality reduction layer to assure approximately equal overall number of parameters in all the three models for a fair comparison. DPGCNN uses primal and dual convolution in every layer. Each dual convolutional layer receives as input for each edge the refined edge features concatenated with the refined vertex features of its incident nodes produced by the previous dual/primal convolutional layer. Edge features were initialized with the features of incident nodes for both Dual GAT and DPGCNN. Mean cross-entropy, dropout with keep probability of 0.9, and learning rate of $10^{-2}$ were used for all models. Table 4.1 presents the link prediction results averaged on 100 runs, showing that DPGCNN outperforms both competitors.

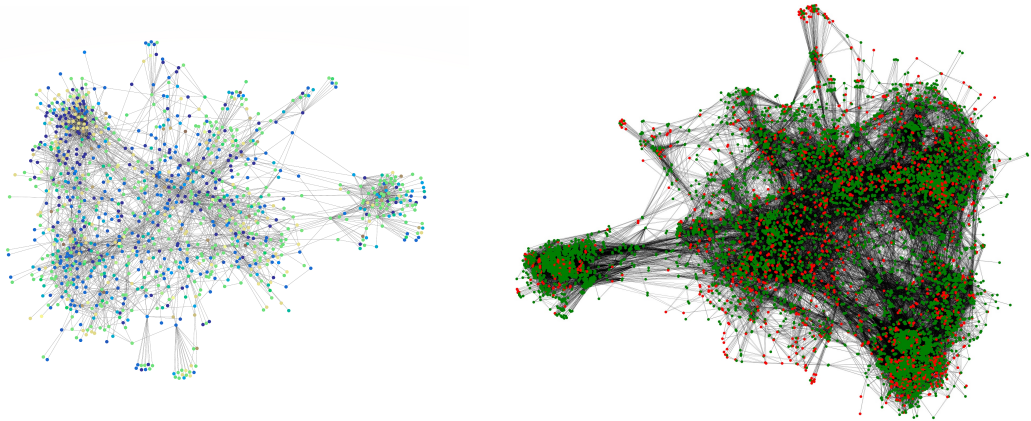

Figure 4: Left: primal graph of the portion of directed CORA used for link prediction. Vertex colors code the ground-truth classes. Right: corresponding dual graph. In green/red are dual vertices (primal edges) that were correctly/wrongly classified, respectively.

### 4.2 Graph-Guided Matrix Completion

In our final experiment, we address the problem of item recommendation, formulated as matrix completion problem on user and item graphs [34]. Such problems are also known under the name of *geometric-* or *graph-guided matrix completion*. The task is, given a sparsely sampled matrix of scores assigned by users (columns) to items (rows), to fill in the missing scores. The similarities between users and items are given in the form of column- and row graphs, respectively. Monti *et al.* [34] approached this problem as learning with a separable recurrent Multi-Graph CNN (MGCNN) architecture, using an extension of ChebNets [10] to matrices defined on multiple graphs in order to extract spatial features from the score matrix; these features are then fed into an RNN producing a sequential estimation of the missing scores. We repeated verbatim the experiment of [34, 27] on several standard datasets used in the recommender systems literature (MovieLens [32], Flixster [23], Douban [30], and YahooMusic [11]), using different convolutional layers inside RMGCNN (Chebyshev [10], Cayley [27], GAT [42] and the proposed Dual-Primal convolution). For reference, we also report the results of some standard matrix completion methods that are not learning-based. For Douban and Yahoo Music datasets, only a single user/items graph was used, as described in [35].

Polynomial filters (Eq. 8) of degree $p = 4$ were used for convolution on the primal graph; different attention scores were computed according to (Eq. 7) for every order and for every diffusion iteration. We used 4 heads on the dual for every dataset besides Douban where just one head has been exploited because of overfitting. Eight features have been produced as output for each head on the dual. GAT hyperparametrs were determined by cross-validation for Flixster and Yahoo Music; the same hyperparameters were used for DPGCNN. For the remaining datasets, we used hyperparameters from [35]. To make the problem more tractable with classic GPUs (in our experiments we used Nvidia Titan X with 12GB RAM), randomly sparsified versions of the dual graph were used with DPGCNN for Movielens (100 neighbors in the dual), Flixster (18 neighbors) and Yahoo Music (30 neighbors). Such sparsification was pre-computed and fixed throughout the entire learning and testing process. Table 4 summarizes the results achieved with different architectures. DPGCNN outperforms all the competing Graph CNN architectures on all the considered datasets, and beats by a significant margin the standard recommendation systems. Table 5 compares the number of parameters required by MGCNN implemented with GAT and DPGCNN.

Table 4: Performance (RMSE) of several matrix completion methods on the MovieLens, Flixster, Douban and Yahoo Music datasets (– indicates that the result was not reported in the original paper). GAT's performance has been computed in this work.

| Method | MovieLens | Flixster | Douban | Yahoo |
|---|---|---|---|---|
| IMC [22, 45] | 1.653 | – | – | – |
| GMC [24] | 0.996 | – | – | – |
| MC [9] | 0.973 | – | – | – |
| GRALS [39] | 0.945 | 1.245 | 0.833 | 38.042 |
| MGCNN Chebyshev [35] | 0.929 | 0.926 | 0.801 | 22.415 |
| MGCNN Cayley [27] | 0.922 | – | – | – |
| MGCNN GAT [42] | 0.929 | 0.931 | 0.791 | 22.102 |
| MGCNN **Dual/Primal** | **0.915** | **0.902** | **0.789** | **21.970** |

Table 5: Number of parameters for MGCNN with GAT layers and our DPGCNN. Note how both solutions present same number of parameters for all considered datasets. The increased number of parameters for Douban and Yahoo is due to the single graph used for implementing MGCNN.

| Method | MovieLens | Flixster | Douban | Yahoo |
|---|---|---|---|---|
| GAT-MGCNN [42] | 23K | 22K | 41K | 41K |
| **Dual/Primal-MGCNN** | 25K | 24K | 42K | 42K |

## 5 Conclusions

We presented DPGCNN, a Dual-Primal Graph Convolutional Network able to realize rich convolutional filters by operating on both the primal and the dual graph. Our architecture achieves state-of-the-art performance on vertex classification, link prediction and matrix completion problems by requiring a small amount of additional parameters. In future works we plan to further investigate the importance of the dual graph by exploring applications to Computer Vision and Graphics (e.g. point clouds) as well as analyzing new and interesting datasets where features are available not only for the nodes of the given graph but also for the provided edges (e.g. citation networks with features describing co-authorship between two citing works).

## Acknowledgments

FM and MB are supported in part by the ERC Consolidator Grant No. 724228 (LEMAN), Google Faculty Research Awards, Amazon AWS Machine Learning Research grant, Nvidia equipment grant, Radcliffe Fellowship at the Institute for Advanced Study, Harvard University, and a Rudolf Diesel Industrial Fellowship at the Institute for Advanced Study, TU Munich.

OS and AB are supported in part by the German Research Foundation, Emmy Noether grant GU 1409/2-1, and by the Technical University of Munich - Institute for Advanced Study, funded by the German Excellence Initiative and the European Union Seventh Framework Programme under grant agreement no 291763, co-funded by the European Union.